\documentclass[12pt,a4papr]{article}
\usepackage{geometry}
\usepackage{natbib}

\usepackage{pdfpages}
\usepackage{import}
\usepackage[hyphens]{url}
\usepackage{microtype}
\usepackage{longtable}
\usepackage{array}
\usepackage{booktabs}
\usepackage{tocbasic}
\usepackage[utf8]{inputenc}
\usepackage[T1]{fontenc}
\usepackage{makeidx}
\usepackage[english]{babel}
\usepackage{graphicx}
\usepackage{float}
\usepackage{caption}
\usepackage{subcaption}
\usepackage{amsmath, amssymb}
\usepackage[linesnumbered,lined,boxed,commentsnumbered]{algorithm2e} 
\usepackage{bm} 
\usepackage{rotating}
\usepackage{color}
\usepackage{listings}
\usepackage[anythingbreaks]{breakurl}
\usepackage{lmodern}
\usepackage{enumitem}
\usepackage{calc}
\usepackage{amsmath, amssymb}
\newcommand*\mean[1]{\overline{#1}} 
\usepackage{chngcntr}
\usepackage{pdflscape}
\usepackage{makecell}
\usepackage{titlesec} 
\usepackage{layouts} 

\newcommand{\code}[1]{\texttt{#1}}

\titlespacing*{\section}
{0pt}{0.7cm}{0.7cm}
\titlespacing*{\subsection}
{0pt}{0.5cm}{0.5cm}

\DeclareMathOperator*{\argmin}{arg\,min}
\DeclareMathOperator*{\argmax}{arg\,max}
\DeclareMathOperator{\E}{\mathbb{E}}
\DeclareMathOperator{\Lset}{\mathcal{L}}

\DeclareMathOperator{\Bset}{\mathcal{B}}

\DeclareMathOperator{\Xset}{\mathcal{X}}

\DeclareMathOperator{\x}{\mathbf{x}}
\DeclareMathOperator{\thetab}{\pmb{\theta}}
\DeclareMathOperator{\w}{\mathbf{w}}
\DeclareMathOperator{\X}{\mathbf{X}}

\DeclareMathOperator{\Var}{\mathrm{Var}}

\DeclareRobustCommand{\1}{\text{\usefont{U}{bbold}{m}{n}1}}
\newcommand{\blanco}[1]{}

\begin{document}

\title{Random Boosting and Random$^2$ Forests - A Random Tree Depth Injection Approach}
\author{Tobias Markus Krabel\thanks{8080 Labs, Hanauer Landstraße 150, 60314 Frankfurt am Main, Germany, \emph{tobias.krabel@8080labs.com}}
\and Thi Ngoc Tien Tran
\thanks{Department of Statistics, TU Dortmund University, Vogelpothsweg 87, 44227 Dortmund, Germany, \emph{tien.tran@tu-dortmund.de}}
\and Andreas Groll
\thanks{Department of Statistics, TU Dortmund University, Vogelpothsweg 87, 44227 Dortmund, Germany, \emph{groll@statistik.tu-dortmund.de}}
\and Daniel Horn
\thanks{Department of Statistics, TU Dortmund University, Vogelpothsweg 87, 44227 Dortmund, Germany, \emph{dhorn@statistik.tu-dortmund.de}}
\and Carsten Jentsch
\thanks{Department of Statistics, TU Dortmund University, Vogelpothsweg 87, 44227 Dortmund, Germany, \emph{jentsch@statistik.tu-dortmund.de}}}

\maketitle

\setlength{\parindent}{0pt}

\setlength{\columnsep}{15pt}

\textbf{Abstract}
The induction of additional randomness in parallel and sequential ensemble methods has proven to be worthwhile in many aspects. In this manuscript, we propose and examine a novel random tree depth injection approach suitable for sequential and parallel tree-based approaches including Boosting and Random Forests. The resulting methods are called {\it Random Boost} and {\it Random$^2$ Forest}. Both approaches serve as valuable extensions to the existing literature on the gradient boosting framework and random forests.
A Monte Carlo simulation, in which tree-shaped data sets with different numbers of final partitions are built, suggests that there are several scenarios where {\it Random Boost} and {\it Random$^2$ Forest} can improve the prediction performance of conventional hierarchical boosting and random forest approaches. The new algorithms appear to be especially successful in cases where there are merely a few high-order interactions in the generated data.  In addition, our simulations suggest that our random tree depth injection approach can improve computation time by up to $40\%$, while at the same time the performance losses in terms of prediction accuracy turn out to be minor or even negligible in most cases.
\\

\textbf{Keywords}:
Ensemble schemes, Boosting, Random Forest, MART, Additional randomness.

\setlength{\parindent}{10pt}

\section{Introduction}\label{sec:intro}

In both parallel and sequential ensemble methods the induction of additional randomness has proven to be worthwhile in various aspects. Motivated by such findings, this work is devoted to a novel random tree depth injection approach to complement the existing literature on sequential and parallel ensemble methods using {\it trees} as base learners. These include e.g.~the {\it gradient boosting} framework introduced by \citet{friedman2001greedy} and the random forests pioneered by \citet{breiman2001random}. Inspired by the \emph{stochastic} gradient boosting framework presented in \citet{friedman2002stochastic}, where randomization is combined with the boosting paradigm, in the present manuscript, we propose an algorithm that fits a sequence of regression trees just like boosting, but injects trees of random depth in a new way. In particular, the method is a modification of the well-known multiple additive regression trees (MART) algorithm proposed by \citet{friedman2001greedy} and is called {\it Random Boost} (RB). In a similar manner, we propose a novel variant of the random forest (RF) algorithm called {\it Random$^2$ Forest} (R2F), which relies on the same random tree depth injection approach. More specifically, each tree in the (sequential or parallel) ensemble is built up to a certain depth that is randomly drawn between one and a pre-specified upper bound $d_{max} \in \mathbb{N}$, which is the central parameter of this extension.

In order to investigate RB's and R2F's potentials, we conduct two simulation experiments.
In the first experiment, we analyse the general capabilities of random depths, hence, MART versus RB and RF versus R2F.
Here, two aspects have to be considered at the same time:
On the one hand, naturally, we are interested in the performance of the resulting models.
On the other hand, it should be faster to fit models with random depths, since in general, the used trees are shorter.
Hence, our simulation study looks at both aspects in a multi-objective way, and we calculate Pareto-fronts for all four methods 
in different data situations.
We will see that the Pareto-front of RB clearly outperforms the front of MART, while the situation is more distinct for R2F and RF: Although R2F is not able to reach as high performing models as RF does, it does have some runtime advantages.
In our second experiment, we have a look at real machine learning experiments, including an extensive hyper-parameter tuning.
Again, we compare two objective: The duration of the entire tuning procedure and  the performance of the final model.
Here we will see that both R2F and RB have a 40\% runtime advantage over their counter parts.
However, while RB is able to reach slightly better performances than MART, R2F is outperformed by RF.
Nevertheless, this makes random tree depth injection a complementing feature that is at least worthwhile to be further explored and tested by researchers and practitioners.

The remainder of the manuscript is structured as follows. Section \ref{sec:theory} sets the theoretical foundations of the proposed {\it Random Boost} and {\it Random$^2$ Forest} methods. We review basic regression tree approaches and summarize the state-of-the art in the research area on sequential and parallel ensemble methods. Next, in Section~\ref{sec:add:random} we introduce the notion of
the additional inclusion of randomness in tree-based methods, resulting in the {\it Random$^2$ Forest} (see Section~\ref{sec:r2f}) and in {\it Random Boosting} (see Section~\ref{sec:ran:boost}). An extensive investigation of RB's and R2F's relative merits is conducted on simulated data in Section~\ref{sec:sim}.
Finally, Section~\ref{sec:concl} concludes and outlines possibilities for further research and development.


\section{Theoretical foundations and state of the art}\label{sec:theory}

In this section, we set the theoretical foundations of the framework of tree-based sequential and parallel ensemble methods, where our proposed {\it Random Boost} (RB) and {\it Random$^2$ Forest} (R2F) methods are embedded in. In particular, we shortly explain both the concepts of {\it regression trees}, {\it gradient boosting} and {\it random forests} and demonstrate the additional value of randomness in such algorithms.

\subsection{Regression trees}\label{sec:reg:trees}
\noindent In general, regression trees belong to a class of models that can be described through a conceptually simple yet powerful recursive decision rule: given the learning set $\Lset$ with the set of predictors $\x = (x_1, \dots, x_p)^T$ and real-valued target $y$, choose the best predictor and corresponding values to split the data into $M$ disjoint partitions or regions $R_1, R_2, \dots, R_M$, such that the partitions are becoming more homogeneous with respect to $y$, i.e.\ such that the high response values are effectively separated from the low values. In order to assess the homogeneity of a partition,  trees usually estimate a simple model (such as a constant or a linear model) in each constructed partition and arrange them in a way the partition-specific models minimize some impurity measure
across all regions. After having obtained the subsets, the partitioning continues until a termination criterion has been reached.

Research literature has brought forward a diversity of implementations of such tree-based rule systems. However, the most prominent regression system, the classification and regression tree (CART) devised by \citet{breiman1984classification} is still most frequently used in machine learning (ML), especially as a base procedure in ensembles such as boosting and random forests.

CART conducts binary splitting, i.e.\ $M = 2$, and fits a simple constant in each created subset of the data, making the model both simple to understand and fast to fit. Generally, the procedure itself contains two steps: first, the tree is grown to its maximally possible size and second, it is then pruned back to the optimal size.  However, when used as a base learner, pruning usually is skipped.

\begin{figure}[t]
    \centering
    \begin{subfigure}[]{0.7\textwidth}
    \centering
        \includegraphics[width=0.8\linewidth]{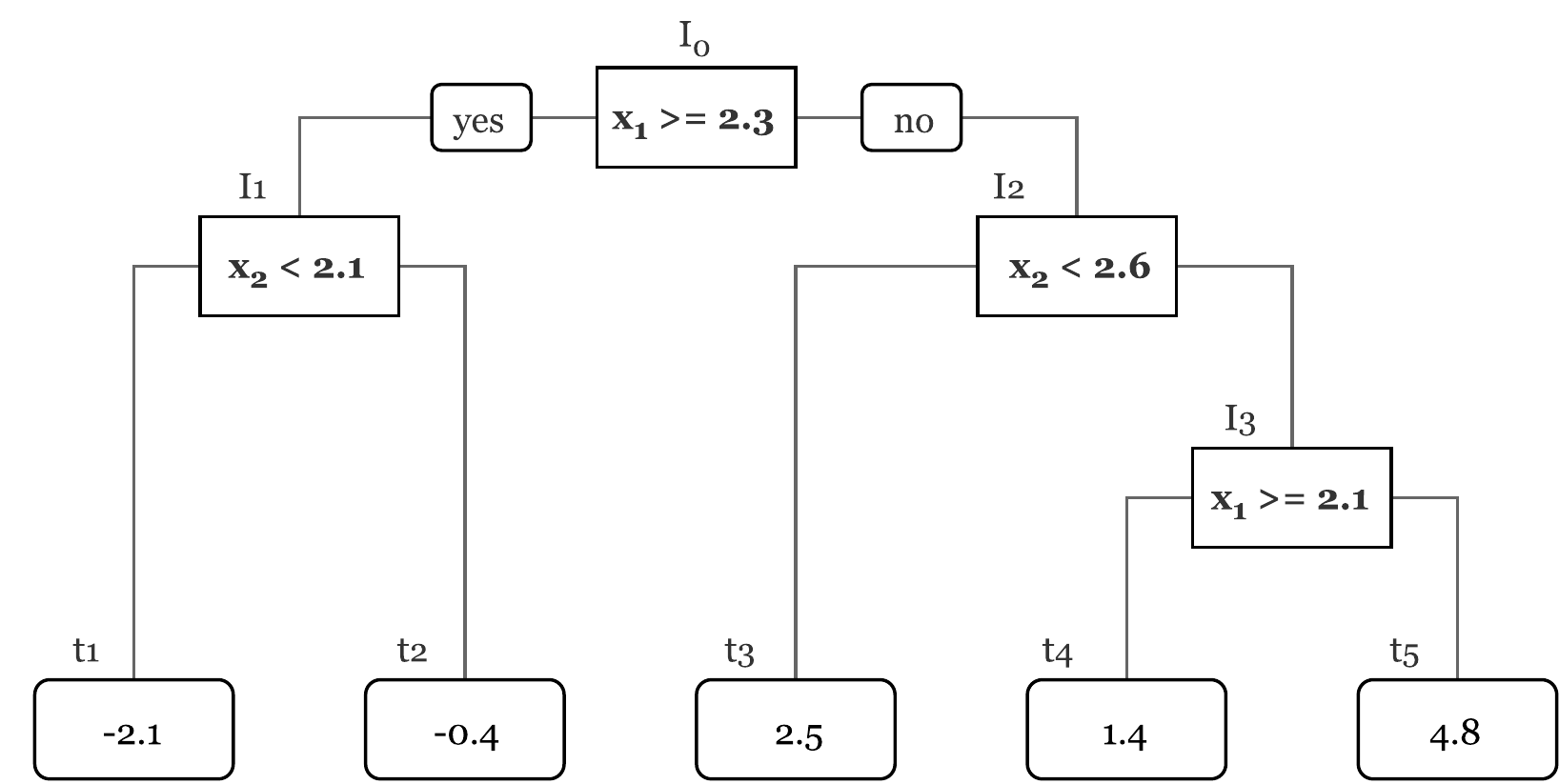}
        \caption{CART structure} \label{fig:cart_structure}
    \end{subfigure}\vspace{0.5cm}

    \begin{subfigure}[]{0.35\textwidth}
        \centering
        \includegraphics[width=0.8\linewidth]{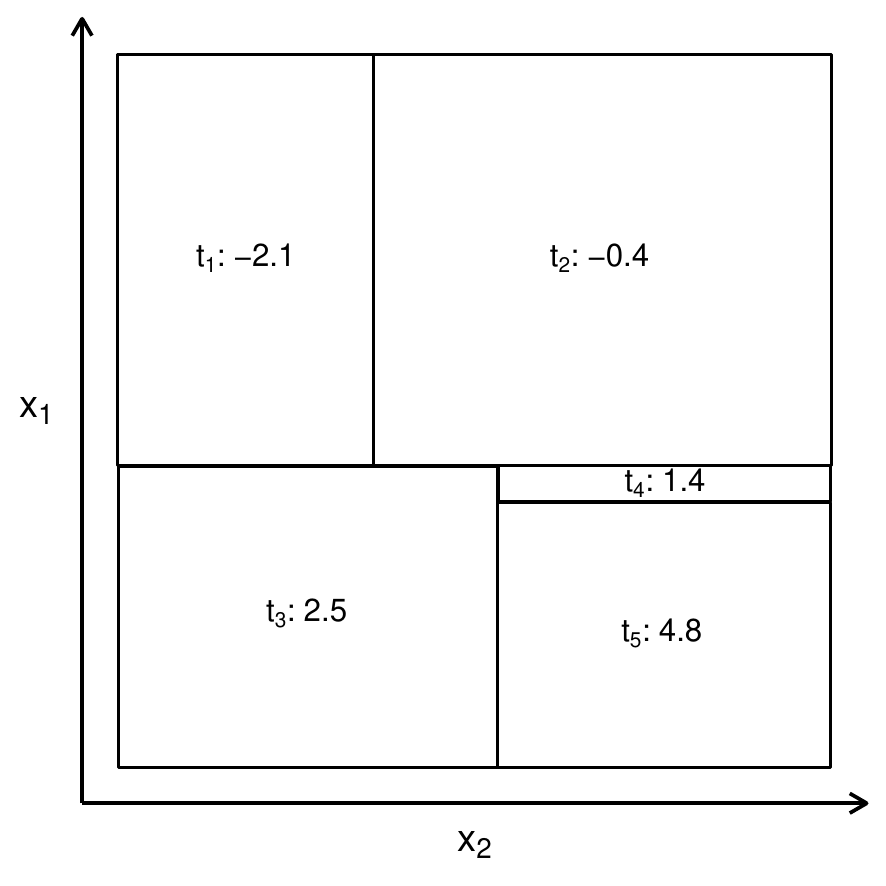}
        \caption{Partitions} \label{fig:2dplot}
    \end{subfigure}
    \begin{subfigure}[]{0.35\textwidth}
        \centering
        \includegraphics[width=0.8\linewidth]{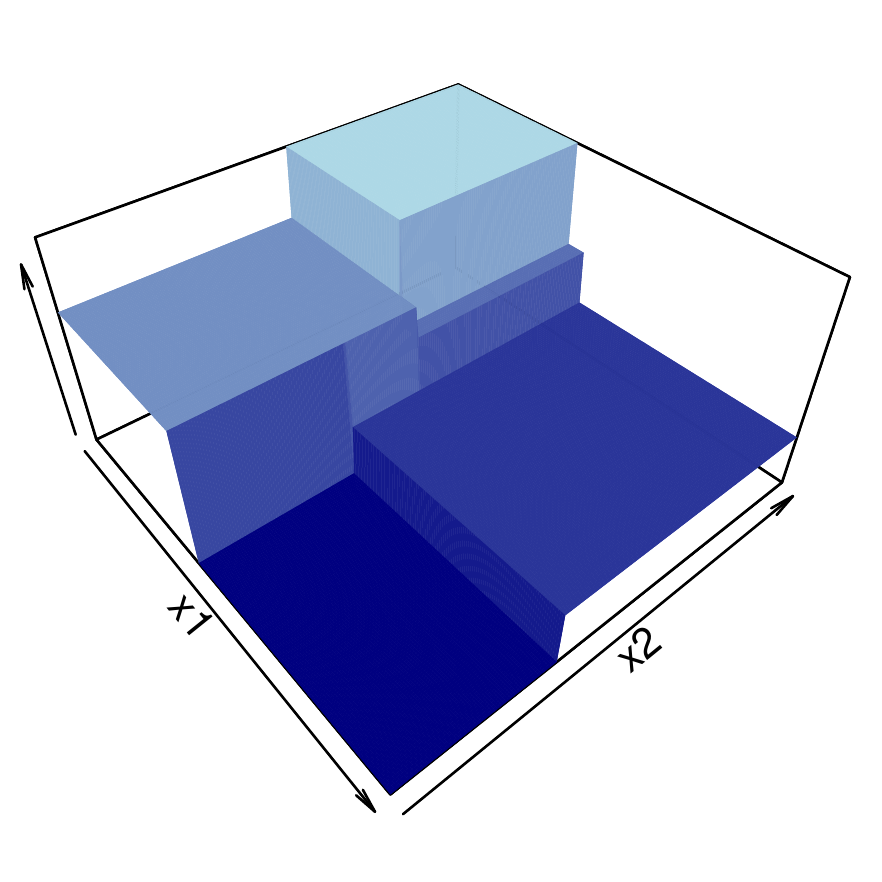}
        \caption{CART prediction surface} \label{fig:3dplot}
    \end{subfigure}
    \caption[CART structure and regression surface]{This figure shows a CART grown on artificial data with feature space $\Xset = [1,4] \times [1,4]$. Panel~(\subref{fig:cart_structure}) shows the tree's structure, Panel~(\subref{fig:2dplot}) displays the partioning of the two-dimensional feature space and Panel~(\subref{fig:3dplot}) the regression surface.} \label{fig:cart}
\end{figure}

Tree growing begins with the full data set $\Lset = \{ (\x_i, y_i) \}_{i = 1}^{N}$, which forms the root node of the tree. Then, a single predictor $x_j$ with corresponding splitting value $s$ is used in order to divide the feature space into a left and right partition/node:
\[
	R_1 = R_1(x_j, s)=\{\x|x_j \le s\} \hspace{0.25cm},\quad\quad \hspace{0.25cm} R_2 = R_2(x_j, s)=\{\x|x_j>s\}.
\]
Within each partition, CART then simply computes the average response value, which determines the corresponding sum of squared errors (SSE).
The objective of CART is to find the split that minimizes the SSE sum of both partitions.
As the variance of values within the parent is always larger than or equal to the sum of the variances of its two children, CART will grow the tree so long as a certain constraint is binding.
After having created the partitions $R_1$ and $R_2$, the algorithm continues with the procedure within each partition unless it contains less than a certain number of observations. When the algorithm is finished, we have a tree-shaped model that is characterized by a set of terminal nodes or \emph{leaves} $\{t_1, \dots, t_T\}$. $T$ is the overall number of leaves and captures the complexity of the tree. Clearly, the larger $T$, the more capacity the model has. In order to obtain a prediction $\hat{f}(\x)$ for a feature vector $\x$, one simply searches the terminal leaf $t_j$ the vector $\x$ falls into and obtains $\hat{f}(\x) = \mean{y}_j$, where $\mean{y}_j$ denotes the average response value from the observations in terminal node $t_j$. In short, the entire model can be expressed in an additive form
\begin{equation}
\hat{f}(\x) = \sum_{t= j}^{T}\1(\x \in t_j)\gamma_j\,,
\end{equation}
with $\gamma_j\in\mathbb{R}, j=1,\ldots,T$.
Since the model $f(\x)$ is constant over each terminal region, the tree can be thought of as a histogram estimate of the regression surface \citep{breiman1984classification}. This becomes clear when imagining a dataset with two features, as shown in Figure~\ref{fig:cart}. The figure is based on an artificial data set with feature space $\Xset = [1, 4] \times [1, 4]$. Panel~(\subref{fig:2dplot}) shows how the space is partitioned into rectangular regions, Panel~(\subref{fig:3dplot}) illustrates the regression surface of the tree. The structure of the regression tree is displayed in Panel~(\subref{fig:cart_structure}). It shows the input variable and split value that divided each internal node (denoted $I_j$), as well as the constant fit to each terminal node $t_j$.

CART is designed to grow as large as possible, resulting in models that
likely overfit the training set and, hence, perform poorly on new data.
Consequently, trees need to be pruned, e.g.\ by allowing only a certain number of splits in total or by only allowing a split if it leads to a statistically significant separation of the output. However, also this strategy could be
to short-sighted, since insignificant splits could be followed by significant ones \citep{friedman2001elements}.
As a result, \citet{breiman1984classification} devised a different strategy for finding an optimal tree: first, carry out the recursive strategy to get a tree as large as possible and second, reduce its size afterwards in a backward step-wise fashion through cost-complexity pruning. The resulting tree candidates are then
evaluated using cross validation (CV) and either the tree with minimal CV error or the smallest tree that lies within one standard deviation of the optimal tree are chosen -- the latter choice aiming at reducing the risk of overfitting and suggested by \citet{breiman1984classification}.

Due to their flexible nature, regression trees in general -- and especially the CART variant -- are a very popular learning system in data mining for practitioners. Their favorable features make them come close to what is called an ``off-the-shelf'' method -- a method that requires neither tuning of an extensive parameter set of the learning procedure nor carefully preprocessed data \citep{friedman2001elements}. Just to name some of its strengths,
categorical independent variables can be naturally treated by trees without the necessity of dummy encoding, trees are able to cope with missing values through surrogate splits and, hence, do not necessarily need imputation of missing values, and are immune to strictly monotonous feature transformations, so scaling the inputs does not need to be considered, as opposed to penalization approaches such as ridge and lasso regression \citep{hoerl1970ridge, tibshirani1996regression}. Furthermore, trees are practically immune to predictor outliers. Regression trees also naturally perform feature selection, so they are (to some extent) resistant to the inclusion of noise.

However, they also come with considerable drawbacks, one being inaccuracy. The simplicity of the model, namely the division of the feature space into rectangular subspaces, can degrade prediction performance, especially when the true underlying function is smooth \citep{kuhn2013applied}. Another issue with trees is their instability, which lies in their hierarchical structure: a split in a parent influences the splits in its children, grand-children, and so forth, so the effect is propagated down to the subsequent nodes. This inherent instability cannot be fully removed using more stable split criteria \citep{breiman1996heuristics, friedman2001elements}. For these reasons, trees as a stand-alone technique are often not competitive with other ML algorithms such as support vector machines or neural nets \citep{geurts2006extremely}. However, when used as a base learner, regression trees can be combined into one model with improved prediction performance.

In the following, different classes of such model combinations will be presented, starting with what is called {\it bagging}.

\subsection{Parallel ensembles}\label{sec:paral:ens}

To address the instability of CART, \citet{breiman1996bagging} came up with
a solution that does not interfere in the tree construction process and improves prediction performance over a single regression tree.
The proposed model uses bootstrapping and aggregating, which led to its name, \emph{bagging}. For bagging, $B$ bootstrap samples $\Bset_b, b=1,\ldots,B,$ are drawn (independently with replacement) from a learning sample $\Lset$ and similar regression trees are grown on each bootstrap replica, leading to a sequence of learners $\{f_b\}_{b=1}^B$. The final model $\hat{f}_{bagging}$ is obtained via the average over all $B$ trees,
\begin{align}\label{eq:bagging}
	\hat{f}_{bagging}(\x) = \frac{1}{B} \sum_{b = 1}^{B} \hat{f}_b(\x).
\end{align}
Intuitively, the argument for bagging is that a single regression tree may misestimate the conditional expectation of $y$ for a given input $\x$, while by averaging the misestimation cancels out. It is this idea that makes bagging superior to a single regression tree, as \citet{breiman1996bagging} finds with a simulation study on real-world and constructed data sets. Overall, the reduction amounts to 20 to 40 percent compared to a single regression tree. However, it is also shown that bagging is not beneficial if the base procedure is already accurate enough.

Bagging has the advantage of averaging over unstable trees, therefore returning stability to the ensemble prediction, which ultimately improves the prediction performance. Compared with other state-of-the-art methods such as boosting, however, bagging seemed to be less effective. This is also due to the fact that the trees in bagging are correlated as the bootstrap samples they are trained on overlap by roughly two thirds on average \citep{friedman2001elements}.


A very powerful modification of bagging that works with trees that are injected with additional randomness in order to reduce pairwise correlation and, therefore, to leverage the variance reducing effect of averaging, was proposed by \citet{breiman2001random} and is based on the {\it Random Forest} (RF) framework.
More formally, consider a set of trees $\left\{f_b\right\}_{b=1}^B$. For each tree $f_b$, there is a corresponding random vector $\Theta_b$ governing its growth. In case of bagging, $\Theta_b$ corresponds to the bootstrap replica of the training set that adds some slight randomness to that tree. With a RF, we let $\Theta_b$ contain all forms of stochasticity during the tree-growing process itself. All random vectors $\left\{\Theta_b\right\}_{b=1}^B$ are independently drawn from the same distribution. Using the bootstrap sample $\Bset_b$ from $\Lset$ together with random vector $\Theta_b$, the tree $f_b = f(\x, \Theta_b)$ is built.
For any given input from the data, the RF estimator is then obtained by averaging:
\begin{align}\label{eq:rf}
	\hat{f}_{rf}^{B} = \frac{1}{B} \sum_{b=1}^{B} \hat{f}(\x; \Theta_{b})\,.
\end{align}
Note the similarity of \eqref{eq:bagging} and \eqref{eq:rf}, where the latter approach incorporates an additional portion of induced randomness.
\citet{friedman2001elements} show that for a given input vector $\x$ the variance of the model is depending on two factors, namely the correlation between any pair of RF trees and the sampling variance of a single random injected tree, i.e.\ $\Var (\hat{f}_{rf}(\x)) = \rho(\x)\sigma^2(\x)$. Since $\rho(\x)\in[0; 1]$, the variance of the ensemble is never greater than the variance of a individual random tree.
However, the amount of randomness $\Theta$ needs to balance two opposing effects. On the one hand, randomness leads to the desired de-correlation of trees, which drives down $\rho(\x)$ and, therefore, $\Var (\hat{f}_{rf}(\x))$. On the other hand, an extra source of randomness inevitably drives up a tree's variance $\sigma^2(\x)$, which increases $\Var (\hat{f}_{rf}(\x))$. In the optimal case, the increased variance of a single regression tree would not be a problem, for one could bring correlation down to zero. Since in practice, however, this is not possible, one has to recognize this trade-off between tree-wise correlation and variance when selecting the degree of stochasticity.

Another important aspect for RF is the relationship between its prediction error and its base learners. \citet{breiman2001random} derives an upper bound for a
RF's prediction error by showing that the prediction error of the RF is never greater than
the expected prediction error of the single random trees it contains.

Altogether, one can derive several practical conclusions. Random forests eventually work through variance reduction, not through affecting the bias. This becomes clear as the ensemble simply averages over a set of base learners, with the effect similar to bagging. Thus, in order to be effective, a RF needs strong base learners, i.e.\ base learners with high capacity and, therefore, low bias. Base learners are single trees injected with randomness $\Theta$, which essentially governs both the variance of a tree as well as its bias. An increase in randomness will decrease the sampling correlation -- from which the ensemble benefits -- but it also increases the bias of a tree (as it imposes restrictions) as well as its variance (through an extra source of randomness). Whereas the latter can be brought down in the ensemble by the number of trees over which to compute the mean, the former critically depends on the choice of randomness. The optimal amount of stochasticity balances these two opposing effects.

Principally, there are manifold ways of designing $\Theta$, yet, the originally designed mechanism seems to have prevailed successfully. It can be described with the following rule, which is based on the idea of \citet{ho1998random}. Before splitting an internal node, randomly select $m_{try} \le p$ predictors and only search for the optimal variable-value pair among those $m_{try}$ variables. The full implementation of RF in regression can be found in Algorithm~\ref{algo:random_forest} in the appendix.
It is this simple addition to bagging that considerably improves the prediction performance of the system over bagging, making it competitive with boosting in many settings, which is one reason why the method today is so popular among researchers and practitioners. Random forests are investigated in a meta study by \citet{fernandez2014do} using 121 data sets and perform among the best methods, together with the support vector machine \citep{cortes1995support}.

Another feature that makes RF so popular is its small number of tuning parameters. Essentially, there are two hyper-parameters in the algorithm: the number of trees fit for the ensemble, $B \in \mathbb{N}$, and the number of randomly selected features available at each split $m_{try} \le p$. In practice, one would typically set $B$ as large as possible, since the procedure is immune to overfitting in this dimension. If computation time needs to be taken into account, one can grow a RF with $B$ trees and pick the smallest subset of trees which works as well as the full ensemble
This is for example facilitated when working with the \code{randomForest} package in R, as it naturally allows accessing predictions of any subset of the model object \citep{liaw2002classification}. Concerning the parameter $m_{try}$, typically ``good'' values are $\lfloor \sqrt{p} \rfloor$ in classification and $\lfloor \frac{p}{3} \rfloor$ in regression \citep{kuhn2013applied}. Yet, the performance of the ensemble is more prone to changes in $m_{try}$ than to changes in $B$, so it is advisable to determine the optimal value of $m_{try}$ through CV or other resampling procedures that split the data into an analysis and assessment dataset \citep{friedman2001elements}.

Although a powerful procedure, RF comes with some caveats. \citet{segal2004machine} examines the standard variant of RF using simulated and real-world data and reveals some of its weaknesses. He states that RF overfits with respect to its tree size and notes that the data sets commonly used for analyzing the procedure could not reveal this phenomenon. In particular, if many noise variables are present, a maximally grown tree with no complexity-regularization will tend to pick noise as a splitting variable, especially since it is only allowed to pick from a random subset of attribute candidates for a split. It is these caveat that we specifically address with our R2F extension proposed in Section~\ref{sec:r2f}. Moreover, when features are strongly correlated also trees
become more correlated as they more likely pick correlated features, leading to a climb in $\rho(\x)$, which works against the RF's central attempt to de-correlate trees.

Although the original RF approach has proven successful, several modifications/extension are in use. One of them is based on purely randomly built trees, i.e.\ on an algorithm that doesn't use any impurity measure at all. Surprisingly, \citet{mingers1989empirical} could show that this method performs just as well as all the original RF. These findings were later caught up by \citet{geurts2006extremely}, who summarized partial and full random tree injection in an algorithm they call \emph{extremely randomized trees} (\emph{Extra-Trees}). Extra-Trees in essence can be regarded as an extreme case of RF, as it is following the same spirit --  building an ensemble of trees that are exposed to stochasticity.
However, the framework grows each tree on the full learning set instead of on bootstrap replicas. Moreover, at each split, $m_{try}$ candidate attributes are selected at random,
but no optimization over the potential split-value takes place. Instead, for each variable, a random split-value is selected from its domain. Then, among the $m_{try}$ split-variable/-value tuples, the optimal one is picked based on some impurity measure. The fully randomized version of Extra-Trees takes this idea one step further. Here, trees are grown based on one randomly chosen split-variable/-value pair per internal node, resulting in trees that are completely independent from the target values of the learning sample. The aggregation of the constructed trees is employed through the arithmetic mean in regression or a majority vote in classification, just as in the other ensemble methods. In a huge simulation study \citet{geurts2006extremely} show
that Extra-Trees are on average $20\%$ faster than RF, while being comparably accurate. Due to its success, Extra-Trees were included in RF implementations, such as in the R package \code{ranger} \citep{wright2017ranger}, where it was added as a split rule variant to the squared-error based split rule used by the classical RF algorithm.


\subsection{Sequential ensembles}\label{sec:sequ:ens}

While in bagging base learners are built in parallel and independently,
in {\it boosting} -- one of the most powerful learning meta-algorithms introduced in recent history -- learners are trained in a sequential fashion. Boosting fits a sequential ensemble where base procedures are grown in a forward stage-wise manner. As a consequence, base procedures are not independent anymore.

Boosting was originally developed for solving classification problems and was later extended to supervised regression. It was based on the principal of weak learnability \citep{schapire1990strength}. By iteratively applying ``weak'' learners -- i.e.\ learners that are just slightly better than a coin toss --
to different aspects of the data, the ensemble combines their output to a strong committee\footnote{This assertion is somewhat outdated, as in many practical applications base learners are already quite complex. This is affirmed by scientific evidence from \citet{bauer1999empirical}, who show that there are situations in which a boosting ensemble's performance and the complexity of its base learners -- trees in this case -- are positively correlated. Furthermore, since boosting has been used in regression, this notion of weak learnability is not as easily applicable anymore.}.
\citet{schapire1990strength} introduced a first simple classification boosting algorithm using only three base learners, and found that the procedure shows a considerable performance gain over the base procedures. A critical assumption of the method to work was that the weak base learner had a fixed error rate.
This assumption was then dropped by \citet{freund1996experiments}, who published a more practically useful \emph{adaptive boosting algorithm} (\emph{AdaBoost}), which constitutes the official pioneer algorithm in boosting. AdaBoost yields potent results by combining an arbitrary number of weak classifiers into an ensemble. Even though AdaBoost is a classification meta-algorithm, it is worth exploring its core, as it gives key insights into the more complex boosting framework that was used in this work.

The strategy of AdaBoost is different from the simple committee explained above as it transforms $\Lset$ in a new way. In essence, AdaBoost repeatedly learns base procedure $f(\x)$ $M$ times on different versions of $\Lset$, leading to a series of $M$ learners $\lbrace f_m(\x)\rbrace_{m=1}^{M}$.  Each $f_m$ learned different aspects of the data, as it got a differently weighted version of the data. The dependency structure between base procedures now originates from the fact that base procedure $f_m$ produces weight vector $\w_{m} \in \mathbb{R}^N$, which gives more weight to observations that were misclassified by $f_m$. The subsequent procedure $f_{m+1}$ therefore puts more focus on learning observations that are harder to predict. In addition, the lower $f_m$'s misclassification error, the more weight it receives in the ensemble. In the end, using all base procedures, the prediction is obtained through majority vote

\[
	\hat{f}_{AdaBoost}(\x) = \argmax_{y \in \lbrace0,1\rbrace} \sum_{m=1}^{M} \alpha_m \1(\hat{f}_m(\x) = y)
\]

\noindent See Algorithm~\ref{algo_adaboost}  in the appendix for a full description in pseudo-code.

AdaBoost reveals two key features of boosting: (1) every learner in the ensemble learns a different aspect of the data by receiving a transformed version of it as input, and (2) the transformation depends on previous learners -- usually, the data is transformed such that subsequent learners focus on cases that were previously less accurately predicted.

The success of AdaBoost made it one of the most influential algorithms in ML and the preferred meta-algorithm among practitioners as it appears to dominate bagging on most problems. Moreover, its popularity created a new research strand focusing on ways to provide enhancements to it \citep{breiman1998arcing, schapire1999improved, wu2008top}. AdaBoost is a pure algorithmic description of an estimation method, and at the time it was developed, there was only a weak link to statistics.  \citet{friedman2000additive} provided that link, stating that AdaBoost actually maximizes the Bernoulli likelihood to fit an additive model of the form
\[
	f_{AdaBoost}(\x) = \sum_{m = 1}^{M} \alpha_{m}\, f_{m}(\x)\,,
\]
which is equivalent to estimating an additive model with an exponential loss $L(y, f(\x)) = \exp(-yf(\x))$.

Further studies on boosting then especially focused on frameworks
that can be used with any loss function both for classification and regression.
\citet{friedman2001greedy}  regarded boosting as an optimization problem with arbitrary loss function, where the addition of each learner to the preliminary ensemble itself is carried out as a minimization problem with squared-error. He was able to develop a boosting framework that can use any differentiable loss function for the optimization, which makes both classification and regression problems solvable. This framework is the basis for one of the most widely used ML methods in practice, the \emph{multiple additive regression trees} (MART) algorithm. MART is also one of the algorithms used in this work. Through the groundbreaking contribution and its popularity, it is worth exploring the framework Friedman proposes in his paper.

As mentioned above, Friedman's approach combines two concepts that lead to a universal and powerful boosting algorithm:  numerical optimization using (1) gradient-descent in (2) function-space. More formally, given some system with random output $y$ and random input vector $\x$, the goal is to find an approximation $\hat{F}$ to the function $F^*$ which maps $\x$ on $y$ with minimal expected loss.
Usually, one restricts $F$ to originate from a parameterized class of functions $F(\x; \thetab)$, with $\thetab$ being a finite vector of parameters that identifies individual functions from the class. For instance, in linear regression, one has the class of linear functions $F(\x; \bm{\beta}) = \sum_{j = 0}^{J} \beta_{j}x_{j}$, and a particularly valued vector $\bm{\beta}_{0}$ would characterize one function $F(\x; \bm{\beta}_0)$ of the family.
In boosting, focus lies on an ensemble model in additive form
\begin{equation} \label{eq:boosting_additive_form}
	F(\x; \bm{\alpha}, \bm{\beta}) = \sum_{m = 0}^{M} \alpha_{m}\,f(\x;  \bm{\beta}_m)\,,
\end{equation}
where $f(\x; \bm{\beta}_m)$ is a simple parameterized function and $\bm{\beta}_m$ is a vector of some finite dimension that characterizes $f(\cdot)$. The vectors $\bm{\alpha}=(\alpha_0,\ldots,\alpha_M)^T$ and $\bm{\beta}=\left((\bm{\beta}_0)^T,\ldots,(\bm{\beta}_M)^T\right)^T$ collect all parameters of the ensemble. Hence, the function optimization problem becomes a parameter optimization problem,
\[
\thetab^* = \argmin_{\thetab} \E_{Y,\X} \left[L(y, F(\x;\thetab))\right]\,,
\]
with $F^*(\x) = F(\x;\thetab^*)$ and a corresponding adequate loss function $L(y, \cdot)$ that penalizes errors in prediction. \citet{friedman2001greedy} added a new perspective to this optimization problem by solving it with gradient descent, a numerical optimization problem that approaches $\thetab^*$  in steps -- starting with an initial guess $\thetab_{0}$ and then updating this guess by adding increments $\lbrace \thetab_1, \thetab_2, \dots, \thetab_M \rbrace$, also referred to as steps. The final estimate of $\thetab^*$ is
\[
	\hat{\thetab} = \sum_{m = 0}^{M} \thetab_m\,.
\]
The increments $\thetab_1, \dots, \thetab_M$ are computed through an
optimal step in the direction of the negative gradient of the loss.
Choosing $M$ sufficiently large, one arrives at a point $\thetab_{M}$ which
qualifies as a (local and not necessarily a global) minimum.
Hence, besides the importance of an adequate first guess $\thetab_{0}$, two other ingredients to the optimization are critical: the number of boosting steps $M$ and the optimal step length in direction of the negative gradient.

In many cases, finding the global solution to \eqref{eq:boosting_additive_form}
is not feasible, so one uses the greedy stagewise approach where one optimizes
each component of the ensemble's parameter collections $\bm{\alpha}$ and $\bm{\beta}$ separately. Hence, for given data $y_i,\x_i,i=1,\ldots,n$, in iteration $m$ we search
\[
	(\alpha_m, \bm{\beta}_m) = \argmin_{(\alpha, \bm{\beta})} \sum_{i = 1}^{n} L\left( y_i, F_{m-1}(\x_i) + \alpha\, f(\x_i; \bm{\beta}) \right)\,.
\]
Here, $F_{m-1}(\x)$, the result of iteration $m-1$, is treated as a constant in this optimization, which reduces the complexity of the problem, as one now only seeks to find the best base learner that can be added to the existing model. The model is then updated,
\[
	F_m(\x) = F_{m-1}(\x) + \alpha_m\, f(\x;\bm{\beta}_m)\,,
\]
so $f(\x;\bm{\beta})$ can be regarded as the ``greedy'' step towards the estimate of $F^*$, under the constraint that this step stems from a parameterized class of functions $f(\x; \bm{\beta})$. The step is called ``greedy'' as it does not necessarily lead to a global minimum. The central idea in \citet{friedman2001greedy} is that one now finds the optimal step by computing the negative gradient of the loss at every observation $\x_i$, 
and then finds the parameterized base learner $f(\x;\bm{\beta}_m)$ that accurately represents that gradient, i.e.\ is most highly correlated with it.
This can be found via least-squares. 
The optimal step size can then be found via line search, which is equal to the weight $\alpha_m$ of the base learner in the ensemble.

In least-squares regression and using an adjusted loss function $\frac{(y - f(\x))^2}{2}$
, the negative gradient becomes $-g(\x_i) = y_i - F_{m-1}(\x_i)$, so each base learner is effectively trained on the current residuals of the preliminary model. This results in
a translation of gradient boosting to the least-squares case, which is the
aforementioned multiple additive regression trees (MART) algorithm, which is summarized
in Algorithm \ref{algo:ls_boost} in the appendix.

Although being a powerful algorithm, Friedman's {\it gradient boosting algorithm}
needs to be treated with more care than e.g.\ Random Forest in terms of hyper-parameter tuning,
since the method is more prone to overfitting \citep{wenxin1999weak, vinayak2015dart}.
This problem is embedded in its nature, namely the fact that base procedures are gradually learned on residuals.

In order to avoid overcapacity, the model can be tuned with respect to the number
of trees in the ensemble, $M \in \mathbb{N}$. Clearly, the larger $M$ the more complex
the model, hence the more likely overfitting. However, there is another and equally
important parameter that can be used in order to make boosting less greedy: the learning
rate $\nu \in (0,1)$, typically small, e.g.\ $\nu=0.1$ or $\nu=0.05$. The learning rate
constrains the fitting procedure through \emph{local shrinkage} of each update to the
ensemble. This local shrinkage is conducted by replacing line 6 in Algorithm~\ref{algo:ls_boost} with
\[
		F_m(\x) = F_{m-1}(\x) + \nu\, \alpha_m\, f(\x;\bm{\beta}_m)\,,
\]
essentially multiplying each update to the ensemble with a constant
value in $(0,1)$. The learning rate can therefore be regarded as regulating
the step length of each boost, and it is found that this can have strong
effects on overfitting \citep{vinayak2015dart}.

\begin{figure}[t]
	\includegraphics[width=\textwidth]{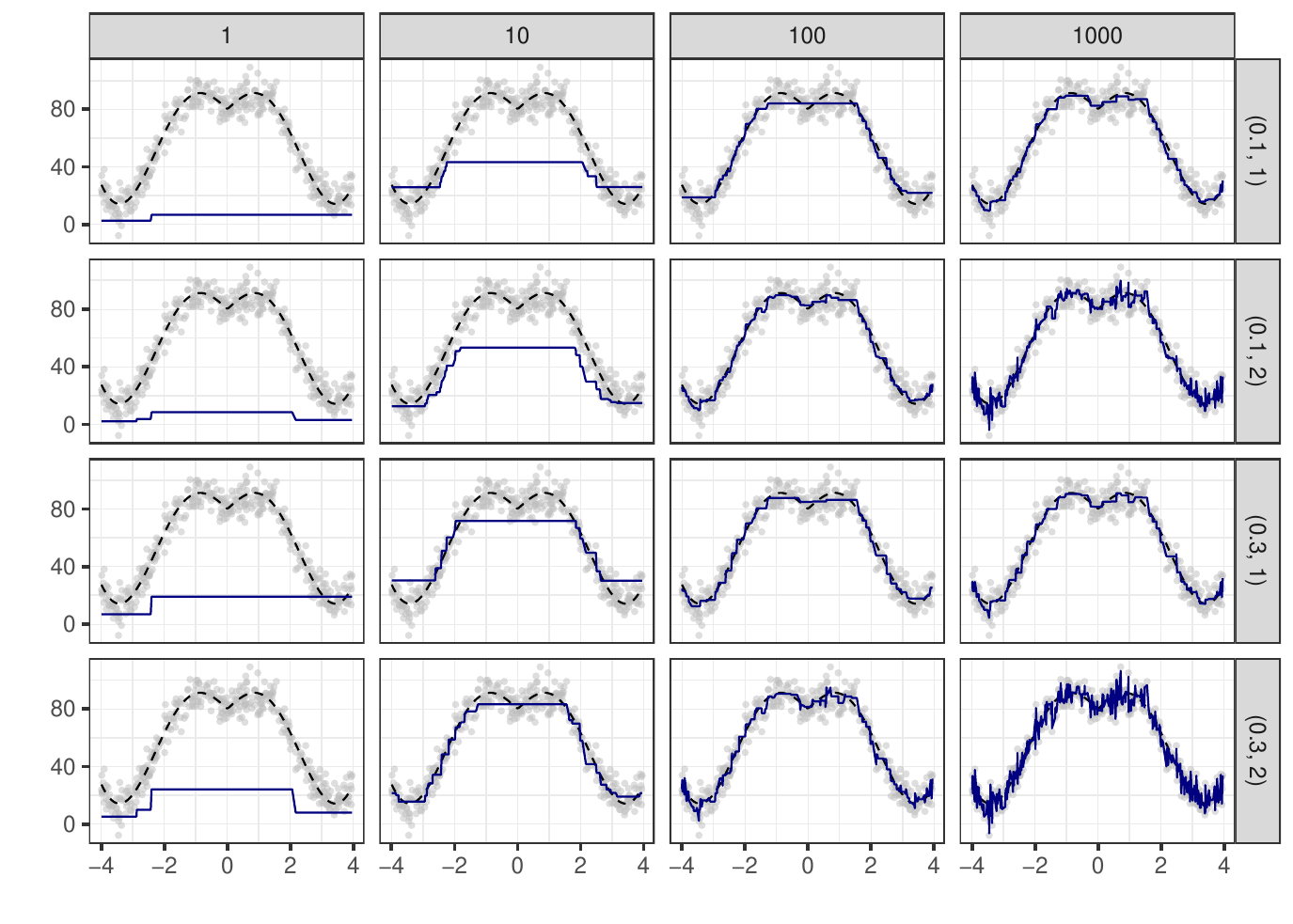}
	\caption[Evolution of fit of MART for varying parameterizations]{This figure
	shows the fits for a MART model for different parameter values. Columns
	refer to the number $M$ of trees the ensemble consists of, rows represent the tuple $(\nu, d_{max})$ containing the learning rate and tree size. The signal is illustrated by the dashed line.}
	\label{fig:mart_fit}
\end{figure}

\begin{figure}[t]
	\includegraphics[width=\textwidth]{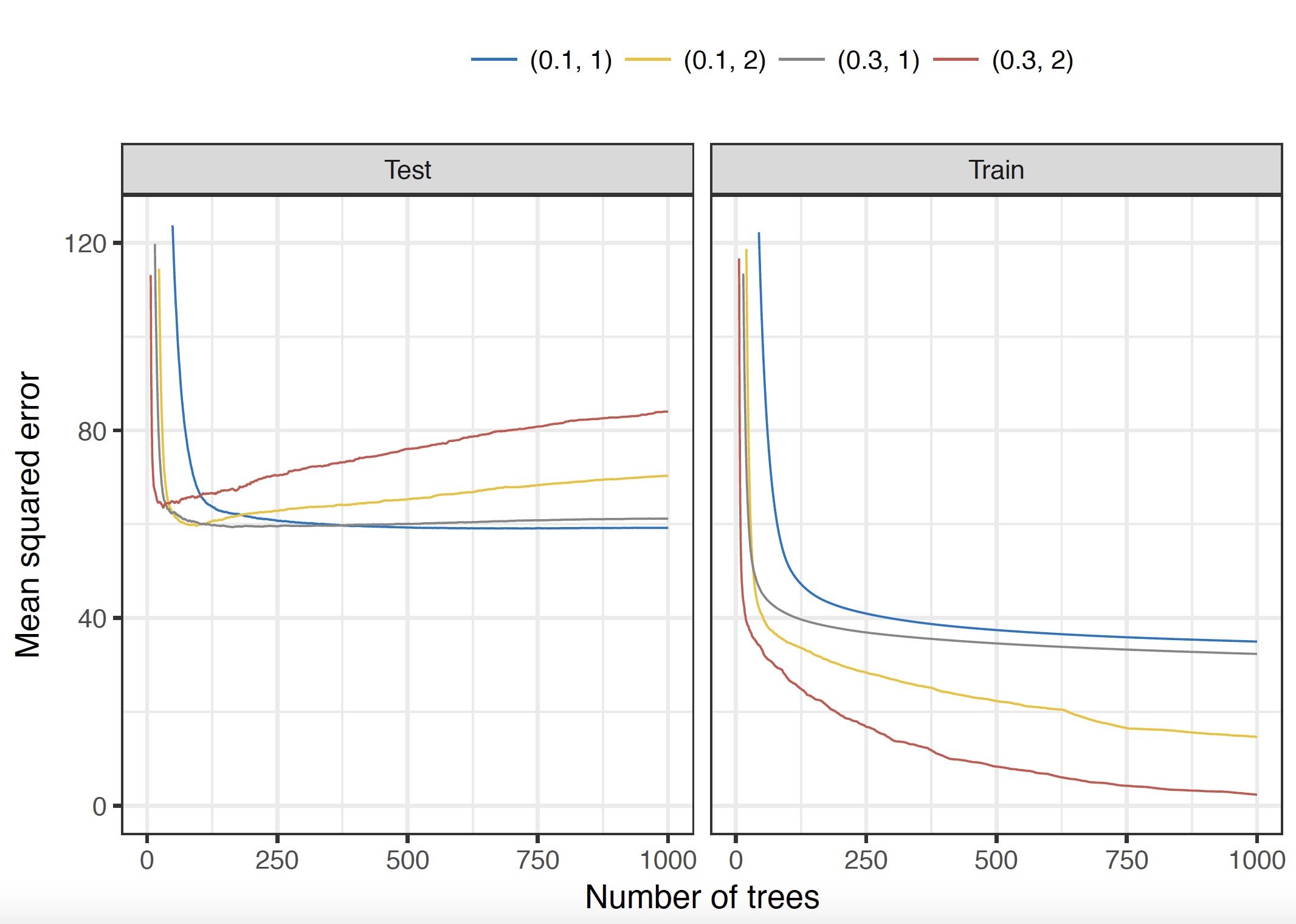}
	\caption[Trajectory of MART train and test error for different parameters]{Test (left)
	and training error (right), respectively, for different parameter constellations of
	MART. Colours represent the tuple $(\nu, d_{max})$, where $\nu$ refers to the learning rate and $d_{max}$ is the maximum size of
	each base learner, i.e.\ each tree can have at most $2^{d_{max}}$ leaves.}
	\label{fig:mart_train_test_error}
\end{figure}

There is a negative relationship between $M$ and $\nu$: The smaller the
learning rate, the more boosts are necessary to achieve an optimal result.
\citet{friedman2001greedy} studied the effect of the two parameters on the
test errors and comes to the conclusion that it is advisable to set $M$ as large
as computationally feasible,
while determining the optimal value for $\nu$
through CV. Another tuning parameter to be considered is the maximum depth
that globally holds for all trees, $d_{max}$, which is correlated with the number
of leaves each tree has and therefore affects the capacity of the base procedures
itself. In order to make the effects of these parameters more tangible, we show
in Figure~\ref{fig:mart_fit} the fits of a boosting model
on some artificial data set following the process
\[
	y = 80 + 20|\x|\cos(\x) + \epsilon, \hspace{0.25cm} \epsilon \sim N(0, 49)\,.
\]
\noindent The columns in Figure~\ref{fig:mart_fit} correspond to $M$, the number of iterations in the ensemble, the rows indicate the tuple $(\nu, d_{max})$. Clearly, one can observe that overfitting increases, ceteris paribus, with an increase in $M$, $d_{max}$, and $\nu$. This finding is supported through Figure~\ref{fig:mart_train_test_error}, which is based on the same data. Here, overfitting is illustrated by showing the training and test error of several MART variants with different parameter values. Whereas the training error for all versions rapidly decreases with the number of trees in the ensemble, the test error starts increasing after reaching a local minimum.

Having introduced both bagging and boosting methods, we finish this section
with a small comparison, which also serves as a short summary of the
discussed methods and as a conclusion of why boosting works. When comparing bagging with boosting from the algorithmic perspective, the differences can be summarized as follows: (1) bagging fits trees in parallel, boosting sequentially, (2) bagging uses resampling, boosting reweighting, (3) bagging does not modify the distribution over cases, boosting does so, and (4) in forming the meta learner, bagging gives equal weights to each weak learner, whereas boosting discriminates between base procedures \citep{freund1996experiments}.

In the next section, we will introduce a modification of MART 
which incorporates another source of randomness into the procedure.
This is done by treating the depths of the single trees as discrete
random variables $D$ with $1\leq D\leq d_{max}$.


\section{Incorporation of additional randomness}\label{sec:add:random}

Having discussed the most well-known state-of-the art ML methods
that are frequently used in practice and are still under the focus of
researchers\footnote{Gradient boosting algorithms are still frequently revised and extended with new features, see, for example, \citet{vinayak2015dart}, \citet{chen2016xgboost} or \citet{thomas2018gradient}.}, it turned out that randomization has shown to be useful in many situations. In the case of bagging, stochasticity is the key to reducing prediction variance, which is the reason why its performance is comparable with boosting in many scenarios. But also in boosting, the idea of incorporating randomization has been utilized to leverage performance.
In 2002, one year after publishing the gradient boosting framework, \citet{friedman2002stochastic} proposed an extension to MART in the form of random subsampling called \emph{stochastic gradient boosting}. Stochastic gradient boosting can be regarded as a hybrid of bagging and boosting where, at each boosting iteration a tree is grown on a random subsample drawn without
replacement from the original training data. The technique has a positive effect on the computation time of the algorithm, as less data need to be processed in order to compute the gradient. Furthermore, \citet{friedman2002stochastic} found that it increases the approximation accuracy of the ensemble. Using a $40$ percent subsample of the data reduces the absolute error by $11$ percent over no subsampling. The fact that this technique works best with high capacity learners suggests that it has a variance reducing effect by de-correlating base learners, just as in RF and Extra-Trees.

Addressing the issue of model over-capacity, \citet{vinayak2015dart} provide a lucrative alternative to local shrinkage. At each boosting step $m$, they compute the gradient of the loss only using a random subset of the model, i.e.\ some trees are dropped -- hence the name \emph{Dropout Additive Regression Trees} (DART). DART relies on a strategy that has already been proposed for learning deep neural nets, where neurons are silenced at random in order to not let the whole net just rely on some few nodes. The number of dropped trees $k$ at each step lies between $0$ and $m$ and makes it behave like MART or RF in the respective extremes, which is why it can be perceived as a mixture of the two models. In an experiment, they show that DART can outperform both MART and RF at a significant margin.

The success of randomization within boosting motivated the two methods developed for this work, which are henceforth referred to as {\it Random$^2$ Forests} (R2F) and \emph{Random Boost} (RB). Both methods are centered around the depth or size of a tree $d \in \mathbb{N}$, which is the maximum number of consecutive nodes in a tree minus one. The size of a tree is the maximally possible interaction order a tree can replicate and is related to the number of leaves $l$ via an exponential relationship. In general, the deeper a tree, the more splits and therefore terminal nodes it \emph{can} have. Note that the regression trees used in all ensembles mentioned above are binary trees. Consequently, a tree with depth 0 can only have one leaf, which is the root, a tree with depth one can have two leaves, a tree with depth two can have four leaves, etc. Thus, the number of leaves of a binary tree is upper bounded by $l \le 2^d$.

\begin{figure}
	\centering
	\includegraphics[width=\textwidth]{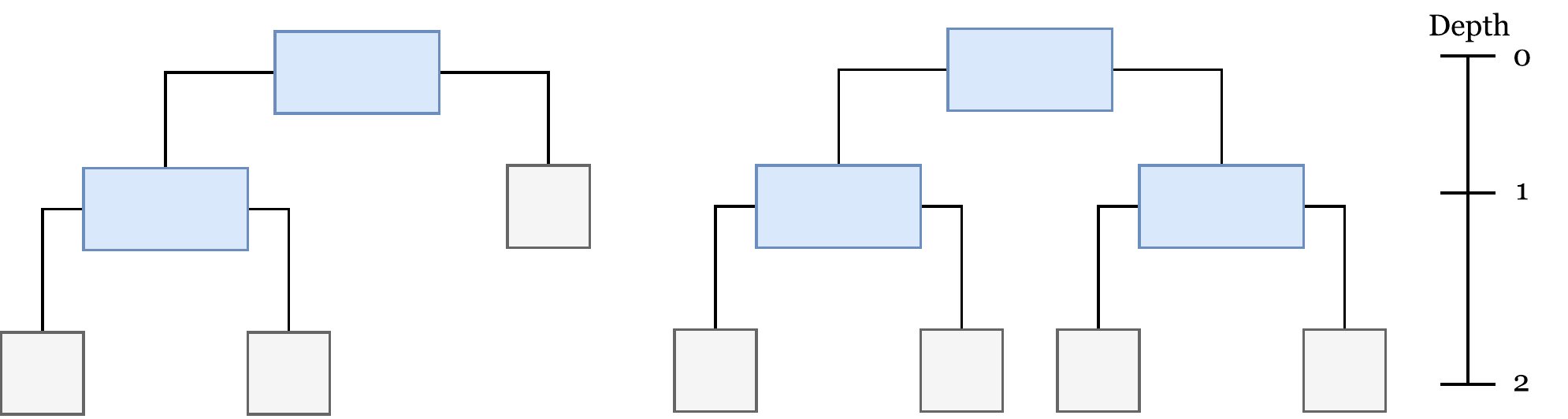}
	\caption[Relationship between tree depth and number of leaves]{Relationship between
	tree depth and the number of leaves using a partially (left) and
	fully grown tree (right) of depth 2 with three and four leaves, respectively (indicated as grey boxes).}
	\label{fig:tree_depth}
\end{figure}

Note that this relation holds in equality if there are enough data points so
that recursion won't stop before the maximally possible number of leaves are created. Otherwise, termination criteria will stop tree construction and the model will not reach its full number of terminal regions, maybe even not its full depth. Figure~\ref{fig:tree_depth} illustrates the relationship between the number of leaves and size with two binary decision trees of depth $2$. Whereas the right tree is fully grown with four leaves, the left one has one split and therefore on terminal node less.

\subsection{Random$^2$ Forests}\label{sec:r2f}

In the following, we now propose an extension to random forests, adding an additional component of stochasticity, which we call {\it Random$^2$ Forests} (R2F). In particular, R2F is new in the sense that for each single tree the actual depth $d$ is now randomly and uniformly\footnote{To the best of our knowledge, this is the first piece of work that experiments with such an additional randomization variant of RF.}.

There are several reasons why R2F can add value to the existing RF procedures. First, R2F is computationally more efficient than RF, when being equipped with the same value for the tree size. If R2F and RF are trained with a value for the maximum tree depth equal to $d_{max}$, then R2F will in many cases grow trees of size $d < d_{max}$ by nature, whereas for RF, all trees will be grown of size $d_{max}$. If we neglect for a moment that trees sometimes are not grown up to their maximum given depth because of a lack of data points in an internal node, then, for RF, the number of splits $s$ for a regression tree of size $d_{max}$ is equal to the number of leaves minus one, $s_{\mathrm{RF}}(d_{max}) = l - 1 \le 2^{d_{max}} - 1$. For R2F, this is different. Given the upper bound, the algorithm grows a tree of size $d$ with
probability $P\left(D = d \right) = \frac{1}{d_{max}}$. Consequently, the expected number of splits in a tree is given by
\[
	s_{\mathrm{RB}}(d_{max}) \le \frac{1}{d_{max}} \sum_{d = 1}^{d_{max}} (2^d - 1)\,.
\]
Assuming equality -- i.e., assuming that during tree construction, no constraints are binding -- and neglecting the constants on both sides, i.e.,
$s_{\mathrm{RF}}(d_{max}) \approx 2^{d_{max}}$ and $s_{\mathrm{R2F}}(d_{max}) \approx \frac{1}{d_{max}} \sum_{d = 1}^{d_{max}} 2^d$, we can derive a closed form approximation of the relative number of splits:
\begin{equation} \label{eq:comptime_prediction}
	s_{rel}(d_{max}) = \frac{s_{\mathrm{R2F}}(d_{max})}{s_{\mathrm{RF}}(d_{max})} \approx \frac{2}{d_{max}}\left(1 - \left(\frac{1}{2}\right)^{d_{max}} \right).
\end{equation}
Further, assuming the computation time for one split is the same for both algorithms and equals some constant, the formula predicts that for size parameter values $d=2$, $d=3$, and $d=4$, R2F takes $75$\%, $58$\%, and $47$\%
of the computation time of RF, respectively. These predictions, however, may
be too optimistic. If trees are not necessarily built to their full sizes, then this arguably decreases computation time stronger for RF than for R2F, as the latter is expected to work with smaller tree depths. Taking this into account should hence bring the two procedures closer together in terms of computation time. Still, the calculations suggest that efficiency gains can be expected.


Second, adding randomness to the tree depth selection adds another source of stochasticity, which further de-correlates trees and, hence, has a variance reducing effect on the ensemble. This is essentially the same argument that motivated \emph{Extra-Trees}, which gives rise to assuming that randomizing over tree depth can have a beneficial effect on the prediction performance, too.

Altogether, the above mentioned arguments make R2F a promising modification of the existing RF framework. The two methods will be compared in detail with regard to goodness-of-fit and computational complexity in simulations in Section~\ref{sec:sim}.


\subsection{Random Boost} \label{sec:ran:boost}

Similar to the R2F from the previous section, also RB is new in the sense that it again treats $d$ as a discrete uniform random
variable\footnote{To the best of our knowledge, this is the first piece of work that experiments with such a randomization variant of boosted trees.}. More specifically, at each iteration, a random tree depth $d$ is drawn between $1$ and some specified upper bound $d_{max} \ge 1$ and a tree is grown based on this information. That is, a sequence of differently sized trees $f_1(\x; \bm{\beta}_1, d_1), \dots, f_M(\x; \bm{\beta}_M, d_M)$ is obtained in a boosting manner and combined in the additive form
\[
	\hat{f}_{RB}(\x) = \sum_{m = 1}^{M} \alpha_{m}\, \hat{f}_m(\x; \bm{\beta}_m, d_m)\,.
\]
Again, similar to R2F from the previous section, there are several reasons why RB can add value to the existing boosting procedures. First, RB is again computationally more efficient than conventional MART, when being equipped with the same value for the tree size for the same arguments as for the comparison of R2F and RF in the previous section. Actually, exactly the same closed form approximation of the relative number of splits given in Equation~\eqref{eq:comptime_prediction} applies.

Second, in the context of boosting two reasons apply to assume that randomizing over tree depths can have a beneficial effect on the prediction performance. From a variance perspective, boosting suffers from overcapacity for various reasons, one being that base learners might be chosen that are too rich in terms of depth. If, for example, one assumes that the dominant interaction in the data generating process is of order three, one would pick a tree with equivalent depth in MART in order to capture this interaction depth. However, this may be overkill, as fully grown trees with depth equal to $3$ have eight leaves and therefore learn noise in the data if there are only a few of such high order interactions. Perhaps, in such a case, a tree with depth $3$ but less than eight leaves would be optimal. This is not accounted for in MART if one does not want to add a pruning step to each boosting iteration at the expense of a computational overhead.
RB may offer a more efficient remedy to this issue. As for the R2F approach, with probability $1 / d_{max}$ a tree is grown
which is able to capture the high order effect at the cost of also learning noise, while in all the other cases,
RB constructs smaller trees that do not show the over-capacity behavior and that can focus on interactions
of smaller order. So it can also be expected that if over-capacity is an issue in MART, e.g.\ due to different
interactions in the data governed by a small number of high order interactions, RB may outperform MART.
Furthermore, similar to the R2F approach, RB also de-correlates
trees trough the extra source of randomness, which has a variance reducing effect on the ensemble.

\emph{This source of randomness, however, bears some risks in this specific context.} Due to the
additive nature of boosting and the potentially resulting path dependencies, RB bears the risk of
very bad predictions if one has bad draws for the depth parameter. As a consequence, the introduced
randomness may overall increase the model's variance, yet with an increasing number of iterations, this
effect should diminish. Altogether, the above mentioned arguments make RB a promising extension to
the existing MART framework. Also these two methods will be compared in detail with regard to goodness-of-fit
and computational complexity through simulations in the next section.


\section{Simulation Studies} \label{sec:sim}

In the following, we present a short simulation study that is based on a data generating mechanism proposed in
\citet{friedman2001greedy}.
Each single data set contains $n$ observations and $p := p_{signal} + p_{noise}$ covariates, however only $p_1$ 
of these variables have a real influence on the response variable $Y$, while the remaining $p_{noise}$ variables are pure noise variables.
The $p$ covariates are generated independently from a standard normal distribution, i.e.\
 $$
 x_{i,1},\ldots,x_{i,p} \overset{i.i.d}{\sim} \mathcal{N}(0,1),i=1,\ldots,n\,.
 $$
The response variable $Y$ has the following structure:
$$
Y_i = F^{*}(\mathbf{x}_{i})+\epsilon_{i}\,,
$$
with covariate vector $\mathbf{x}_{i}=(x_{i,1},\ldots,x_{i,p_{signal}})^T$ and $\epsilon_{i}$ defined below. The function $F^{*}(\cdot)$ is given by
$$
F^{*}(\mathbf{x}_{i}) = \sum\limits_{j=1}^{p_{signal}} a_j\, g_j(\mathbf{z}_j)\,,
$$
where the regression coefficients $a_j$ are uniformly distributed on the interval $[-1,1]$, i.e.\  $a_j\sim \mathcal{U}[-1,1]$, $j=1,\ldots,10$,
and $g_j(\mathbf{z}_j)$ are $n_j$-dimensional Gau{\ss}-functions of the following form:
$$
g_j(\mathbf{z}_j) = \exp\left(  \frac{1}{2}\left[ (\mathbf{z}_j - \pmb{\mu}_{j})^T \mathbf{V}_j (\mathbf{z}_j - \pmb{\mu}_{j})\right]\right)\,.
$$
We have $\pmb{\mu}_{j}\sim\mathcal{N}_{n_j}(\mathbf{0},\text{diag}(1,\ldots,1))$, $\mathbf{z}_j = \{x_{i,P_j(l)}\},$ $l=1,\ldots,n_j$ and $P_j(l)$ is a permutation from the set $\{1,\ldots,p\}$ of natural numbers. Moreover, we define $n_j=\lfloor 1.5+r\rfloor,$ $r\sim \text{Exp}(2)$.
As $\mathbf{z}_j$ is a permuted subset of $\mathbf{x}_j$, for each generated ``Friedman'' data set, the selection of influential
covariates is random.

Furthermore, $\mathbf{V}_j$ is a $(n_j \times n_j)$ dimensional, randomly generated covariance matrix:
$$
\mathbf{V}_j = \mathbf{U}_j \mathbf{D}_j \mathbf{U}_j^T\,,
$$
where $\mathbf{U}_j$ is a random orthonormal matrix and $\mathbf{D}_j=\text{diag}(d_{1j},\ldots,d_{n_jj})$ a diagonal
matrix with random diagonal elements $d_{lj}$ generated by $\sqrt{d_{lj}}\sim\mathcal{U}[0.1, 2]$, $l=1,\ldots,n_j$.

Finally, the residual are drawn from
$$
\epsilon_i\sim\mathcal{N}\left(0, \vert F^*(\mathbf{x}_i) - \text{median}\left(F^*(\mathbf{x}_i)\right) \vert  \right),\:i=1,\ldots,n\,,
$$
hence, the residuals are (independently) normally distributed with mean $\mu_i = 0$ and variance $\sigma^2_i =  \vert F^*(\mathbf{x}_i) - \text{median} \left(F^*(\mathbf{x}_i)\right) \vert$, see (\citealp{friedman2001greedy}, p.\ 1207-1209).


We are conducing two experiments.
In both of them the standard variants of a random forest and a boosting are compared to their variants with random depths, the RB and the R2F respectively.
In the first experiments, we perform a multi-objective optimization of the two objective MSE and training time in order to analyse the general capabilities of the methods.
In the second experiment, we will look at a real tuning situation.
Here, the total runtime of the tuning and the MSE of the best found parameter settings are compared.

\subsection{Multi-Objective Comparison}

In this first experiment, we consider four different Friedman-datasets with 10\,000 observations and $p_{signal}=10$ covariates.
Moreover, each dataset is used with 0, 10 and with 20 noise-covariates, i.e.\ $p_{noise}\in\{0,10,20\}$ covariates that per construction
do not have an influence on the target variable.
Hence, a total of twelve data situations is considered.

For each data situation, we perform a multi-objective parameter tuning.
In multi-objective parameter tuning, the hyper-parameters of the learning methods are optimized not only with respect to one, but with respect to multiple objective functions.
Often, two-dimensional parameter tuning is used in order to analyse the performance-runtime trade-off of certain methods
\citep{horn2017b}.
The goal of this analysis is to understand the capabilities of the different algorithms in the different situations.

This is especially useful here, since we do not only expect the random depth methods
to show better performance than their normal counterparts, but also to be faster.
Since these two objectives interact with each other (typically, a forest / boosting with
more iterations reaches a better performance, but also has a longer runtime), they should be considered at the same time.
Hence, we are interested in calculating the so-called {\it Pareto-front}, which shows the set of optimal trade-offs for the two objectives.
For calculation of the Pareto-fronts we use a multi-objective evoluationary algorithm, the \cite{deb2002}, with 10 generations, 80 individuals per generations and default-settings besides.
All calculations were performed in R \citep{R:2020}, the implementation of the MART algorithms is taken from Python \citep{pyt:2020}.

As for the performance, we calculate the mean squared error of the model
on 5\,000 independent test observations, for the runtime we simply stopped the duration of each call to the training-function.

\begin{figure}[t]
\centering
\includegraphics[width=0.94\textwidth]{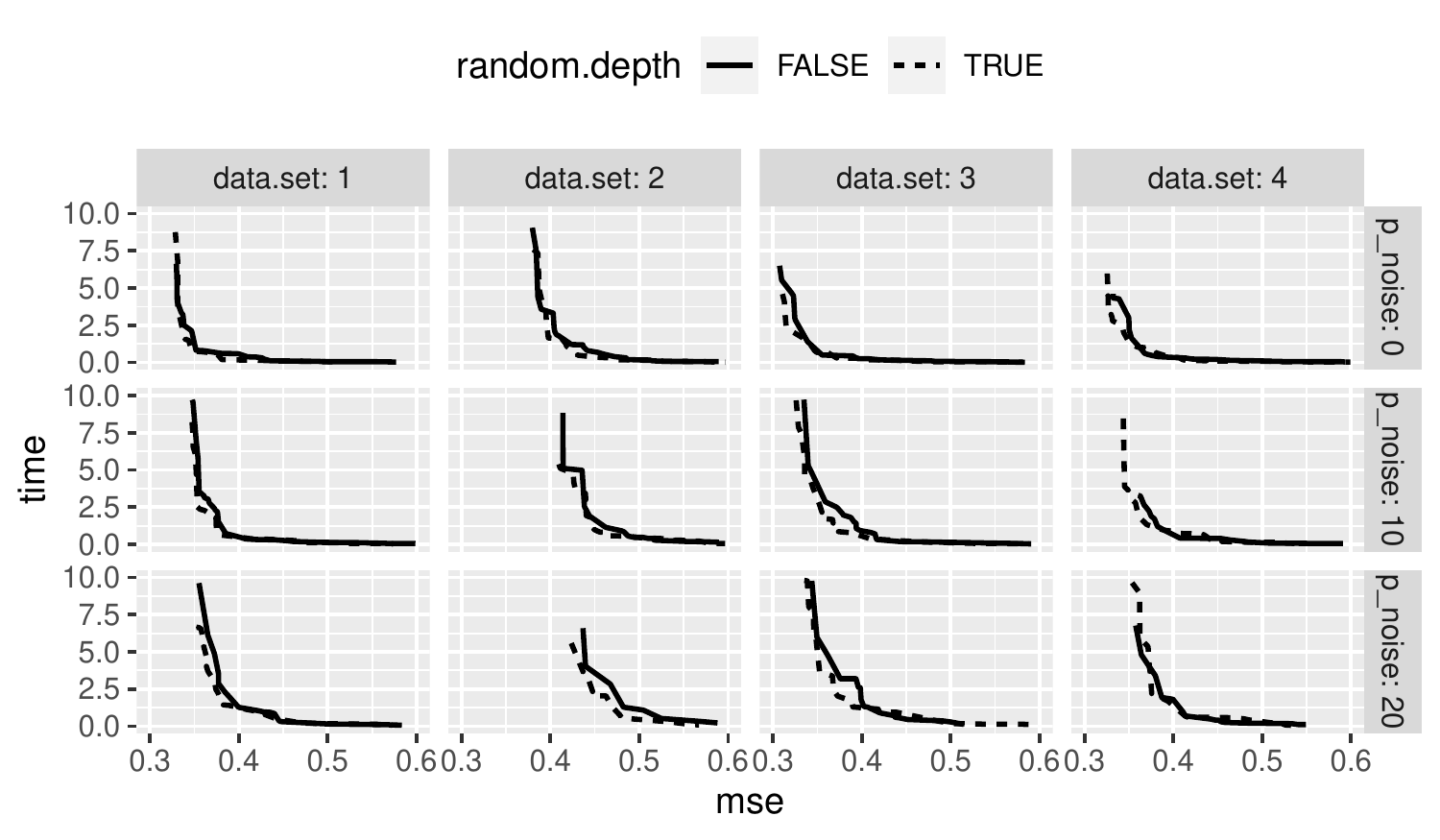}
\caption{Effect of random depth on boosting: Shown are the final Pareto-fronts of multi-objective parameter tunings on 4 data-sets, with $p_{noise}\in\{0, 10, 20\}$ additional noise covariates, respectively.}
\label{fig:fronts_boosting}
\end{figure}

For the boosting, we optimized the following parameters: The number of boosting iterations $m\in \{1, 2, ..., 1\,000\}$,
the learning rate $\nu\in [0, 1]$, the (relative) number of observations used per tree $\lambda\in [0, 1]$ and the (relative) number of variables used per tree $\kappa\in[0, 1]$. The tuning was repeated two times in each
data situation: One time with the random depth enabled (RB), and one time with the random depth disabled (standard MART).
The resulting Pareto-fronts are displayed in Figure~\ref{fig:fronts_boosting} for four exemplary data sets created by the data generating mechamism proposed by \citet{friedman2001greedy} and explained above.

Across all four datasets it can be seen that the RB variant outperforms the default MART algorithm:
The dashed Pareto-front can nearly always be found on the bottom-left side of the solid one and, hence, the RB algorithm is able to reach the same error values as the MART algorithm, but in a lower runtime.
The advantage in runtime is not huge, but it is present throughout.

Moreover, the RB algorithm is able to reach on-par best MSE values as the MART algorithm.
In Table~\ref{tab:fronts_boosting_best_mse}, the differences in the best MSE values are shown, 
here the best value of RB is subtracted from the best value of MART (hence, positive values indicate an advantage for RB).
All differences are rather small, and most times, RB reaches even better MSE values than MART.

\begin{table}[b]
\center
\begin{tabular}{l||r|r|r|r}
dataset &          1  &     2   &    3    &  4 \\ \hline
$p_{noise}$ =  \phantom{0}0 & 0.00002 & -0.00078 & -0.00324 & 0.00665 \\ \hline
$p_{noise}$ = 10 & 0.00726 &  0.00661 &  0.00646 & 0.01342 \\ \hline
$p_{noise}$ = 20 & 0.00370 &  0.00941 & -0.00188 & 0.00567 \\
\end{tabular}
\caption{Difference of best MSE values of RB and MART in Figure~\ref{fig:fronts_boosting}, positive values indicate an advantage of RB.}
\label{tab:fronts_boosting_best_mse}
\end{table}

\begin{figure}[t]
\centering
\includegraphics[width=0.94\textwidth]{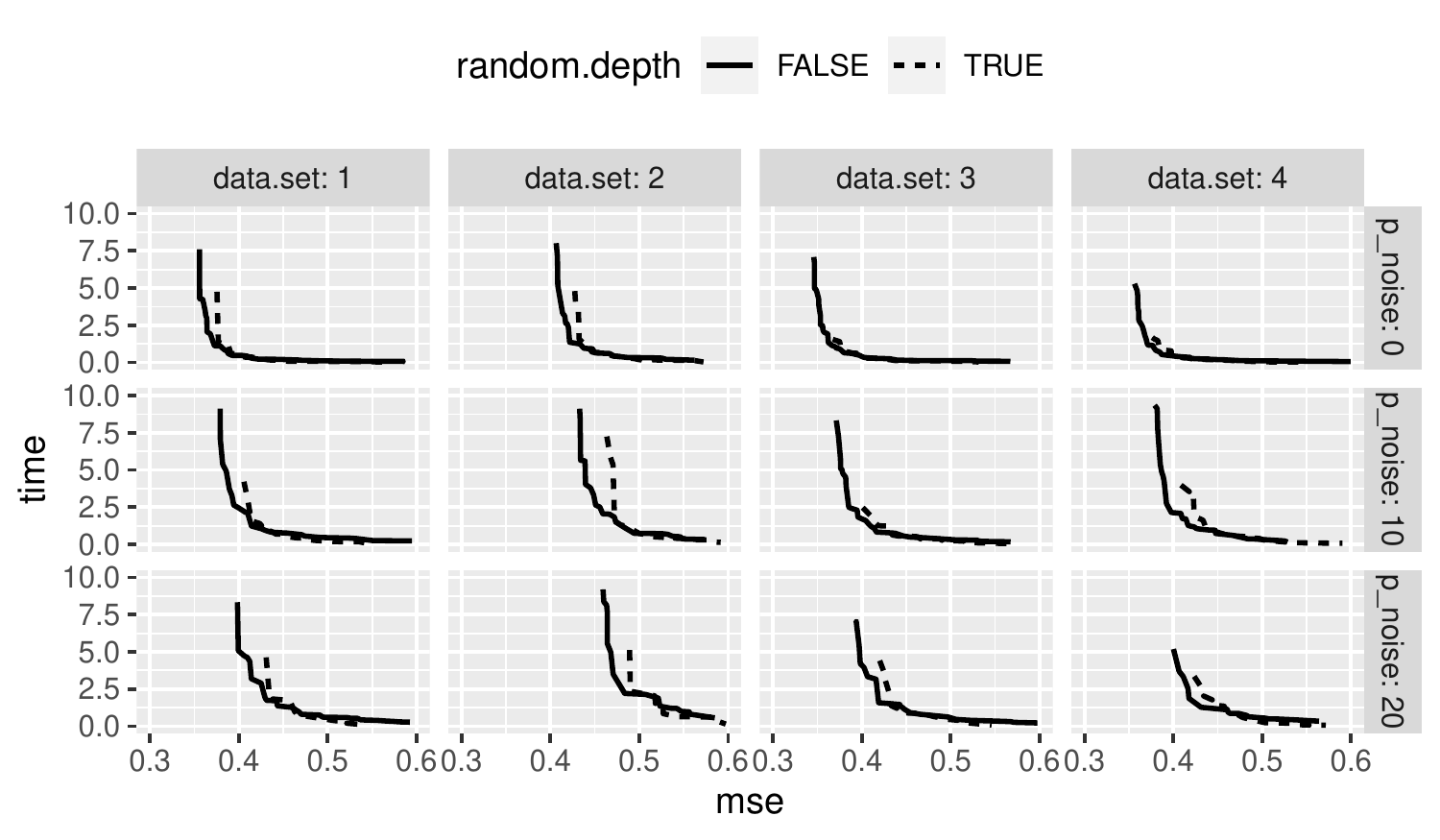}
\caption{Effect of random depth on random forests: Shown are the final Pareto-fronts of multi-objective parameter tunings on 4 data-sets, with 0, 10 and 20 additional noise covariates respectively.}
\label{fig:fronts_forest}
\end{figure}

For the random forest, we optimized the following parameters: The number of trees $n_{tree}\in \{1, 2, ..., 1\,000\}$, the (relative) number of observations used per tree $\lambda\in [0, 1]$, whether observations are sampled with or without replacement and the (relative) number of variables used per tree $\kappa\in[0, 1]$.
Again, the tuning was repeated two times in each data situation: One time with the random depth enabled (R2F), and one time with the random depth disabled (RF).
The resulting Pareto-fronts are displayed in Figure~\ref{fig:fronts_forest}.

Here, the results have to be viewed in a more differentiated way than with boosting, since there is no clear winner along the Pareto fronts.
On the one hand, the front of the RF clearly outperforms R2F in cases with high runtimes and low MSE values.
In particular, the R2F does not appear to be able to achieve models of the same quality as the normal RF.
This can also be seen in Table~\ref{tab:fronts_forest_best_mse}, in every situation the RF reaches clearly better
optimal MSE values than the R2F.

\begin{table}[b]
\center
\begin{tabular}{r||r|r|r|r}
dataset &          1  &     2   &    3    &  4 \\ \hline
$p_{noise}$ =  \phantom{0}0 & -0.02198 & -0.02356 & -0.02364 & -0.02281 \\ \hline
$p_{noise}$ = 10 & -0.03117 & -0.03441 & -0.03311 & -0.03378 \\ \hline
$p_{noise}$ = 20 & -0.04040 & -0.03807 & -0.03431 & -0.03386 \\
\end{tabular}
\caption{Difference of best MSE values of R2F and R2F in Figure~\ref{fig:fronts_boosting}, positive values indicate an advantage of R2F.}
\label{tab:fronts_forest_best_mse}
\end{table}

However, for parameter settings with higher MSE values and runtimes, the R2F is not only able
to be on-par with the RF, but also to be a bit faster.
Now one might think, what the point of being a bit faster with high MSE values is, since,
in the end we are only interested in models with the best performance values.
However, if datasets grow too large, one might not be able to calculate the high-performing
models (since it simply might take too long), hence, the best model given a certain runtime has to be used.
If a model now is able to reach a better performance in the same runtime, it might be preferable.

Moreover, in order to find the best models, a parameter tuning has to be performed.
During this tuning, many different parameter settings are evaluated.
If a method is able to speed up at the evaluation of at least some parameter situations, without losing performance, it might significantly shorten the time needed for parameter tuning.
Hence, we still consider R2F to be promising and look at a realistic tuning situation next.

\subsection{Real Tuning Performance}

In this second experiment, we consider 50 different Friedman datasets with 10\,000 observations each.
Since the results in the first experiment did not show many differences with respect to the number of noise covariates,
we only consider the situation without noise covariates here.

On each of the 50 datasets a realistic parameter tuning using random search is performed.
The performance (i.e.\ the MSE) of 50 different parameter settings is estimated on the
training data set using a 5-fold subsampling: 
the model is learned on 2/3 of the data set, the last third is used to evaluate the model.
This is repeated 5 times, and the mean value of the 5 subsample iterations is used as the performance estimator.
Finally, a model using the best of the 50 evaluated parameter settings is trained on the entire dataset and returned as the tuning result.

The tuning is evaluated in two ways:
At first, an external test set consisting of additional 5\,000 observations is used to estimate the MSE of the final model, second the runtime of the entire tuning process is stopped.
The tuning is repeated both with the random depth enabled and disabled, and, again, we calculate the difference in performance for these two variants on each dataset.

For the boosting, the learning rate $\nu\in [0, 1]$, the (relative) number of observations used per tree $\lambda \in [0, 1]$ and the (relative) number of variables used per tree $\kappa\in[0, 1]$ were optimized, while for the random forest the (relative) number of observations used per tree $\lambda\in [0, 1]$, whether observations are sampled with or without replacement and the (relative) number of variables used per tree $\kappa \in[0, 1]$ were tuned.
The number of trees was set to a constant value $m = 200$ and  $n_{tree}=200$, respectively.

\begin{figure}[t]
\centering
\includegraphics[width=0.9\textwidth]{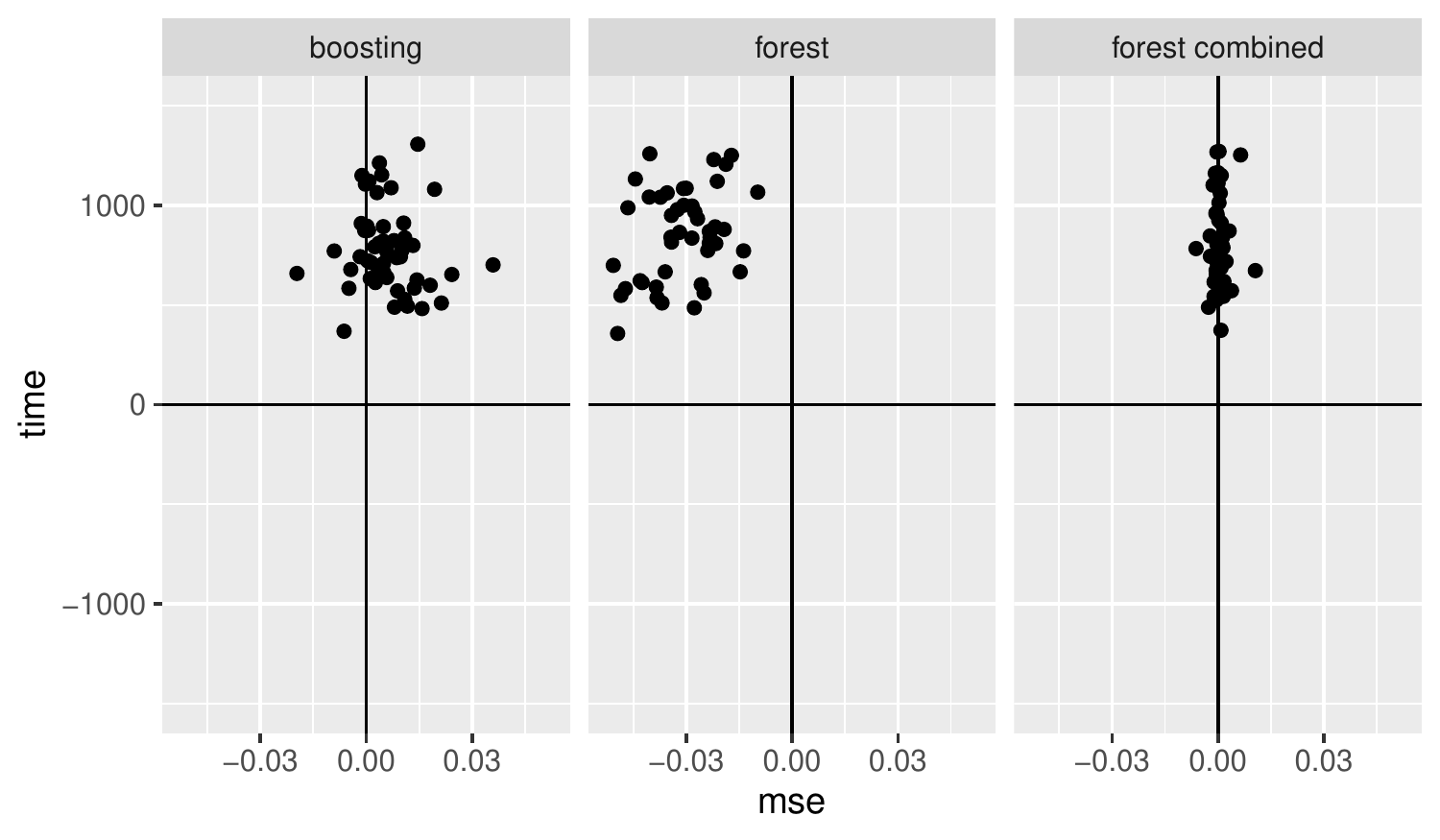}
\caption{Results of the tuning experiments: Scatterplots of the MSE of the final model and the total runtime of the tuning process. Displayed are differences between the models with random depth enabled and disabled (positive values indicate that enabling the random depth reached better results). The left plots shows the results for the boosting-models, the plot in the middle those for the random forest. In the right plot a variant of the random forest is shown, where the random depth was used during the tuning, but not for the final model fit.}
\label{fig:tuning_plot}
\end{figure}

\begin{figure}[t]
\centering
\includegraphics[width=0.9\textwidth]{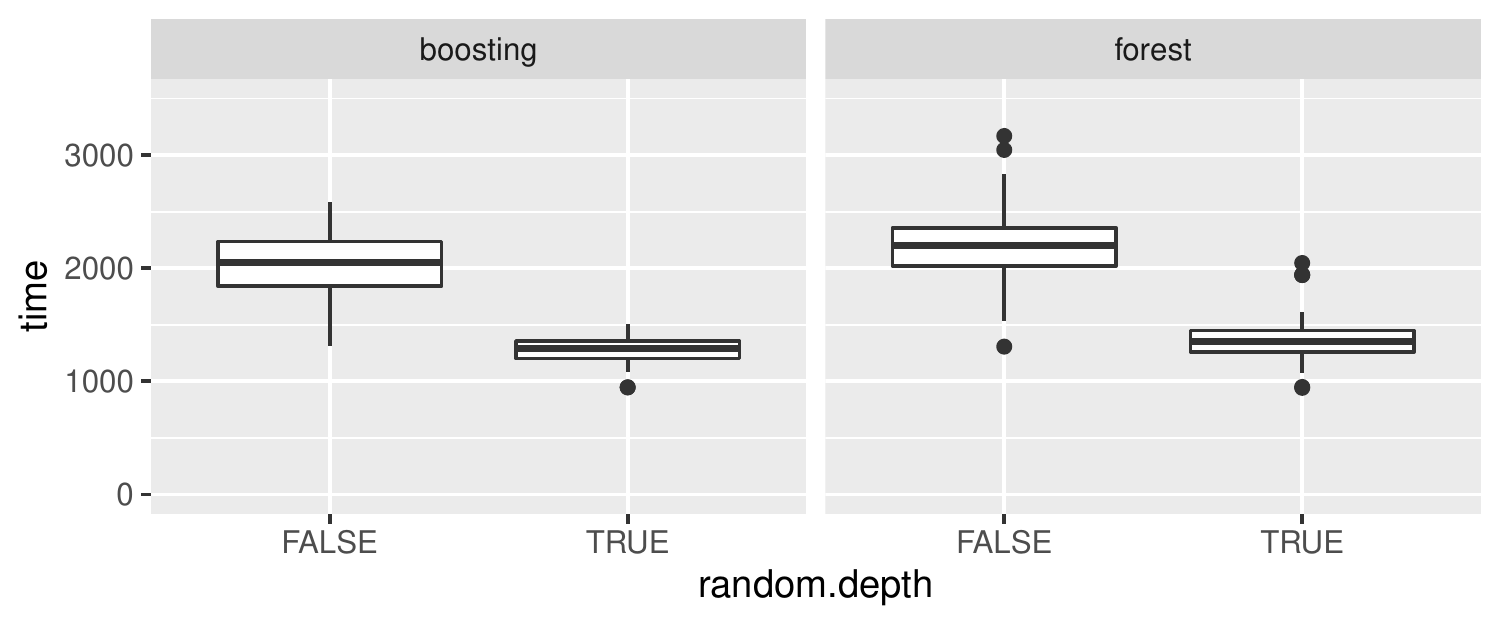}
\caption{Individual runtimes of the parameter tunings.}
\label{fig:tuning_runtimes}
\end{figure}

In Figure~\ref{fig:tuning_plot}, scatterplots of the resulting performance values are shown.
It can be clearly seen that for both the boosting and the forest approaches,
the runtime of the tuning is substantially reduced when enabling the random depth.
For both methods, the tuning with enabled random depth only needed around 60\% of the time the tuning with disabled random depth needed (compare Figure~\ref{fig:tuning_runtimes}).
For the boosting algorithms, RB outperforms MART in most iterations by a bit, or performs just slightly worse.
Hence, with an lower runtime and comparable, perhaps even superior performance, RB is a 
very promising approach to improve existing boosting algorithms.
For the forest approaches, however, the performance of the conventional RF models without the random depths is clearly better than for R2F.

In order to combine the strength of R2F (lower runtime) and RF (better performance), we ran a third experiment:
The tuning is now performed with random depth enabled (hence, R2F is used).
However, as the final model, random depth is disabled and a normal random forest is fitted.
In this way, the tuning can profit from the faster runtime of the R2F, while the final model benefits from the higher performance of the RF.
The assumption behind this idea is that the optimal hyper-parameters of R2F and RF are pretty similar, such that the RF will also achieve good results using the optimal R2F hyper-parameters.
The corresponding results can be found in the right most plot of Figure~\ref{fig:tuning_plot}.
Here it can be seen that the runtime advantage is the same as in the last experiment
(this was to be expected, since the tuning is the same as before).
However, the final performance values are nearly the same, hence, there is no big difference in using the optimal hyper-parameters found using the RF and the R2F.
The resulting method has the same performance as a normal RF, but reaches it in a tuning with a 40\% lower total runtime.


%


\section{Conclusion} \label{sec:concl}

Inspired by theoretical findings on randomization techniques in all different kinds
of ensembles, a new algorithm called Random Boost (RB) was introduced.
The new method that grows trees of random sizes, appears to be a lucrative
alternative to the MART algorithm developed by \citet{friedman2001greedy}
in terms of computation speed and accuracy.
Analog considerations on random forests lead to a corresponding new algorithm Random$^2$ Forest (R2F).

In a multi-objective parameter tuning, both new methods were compared with their counterparts with respect to the objectives accuracy and computation speed.
The results showed that in addition to RB's predictive competitiveness, the algorithm also offers some computation speed improvements worth considering.
Compared to an equivalently parameterized MART, RB majorly grows trees that do not have the full specified depth.
Since this depth is correlated with computation time, RB has an edge over MART in terms of efficiency.
R2F also has an edge in terms of computation time over normal random forests, however, it does fall back in terms accuracy.
Nevertheless, incorporating the new randomization technique into both model types seems promising.

Especially in times of Big Data, where computation time often becomes the bottleneck, every speed up without (too much) loss of performance is very welcome.
It has been shown in a realistic tuning session (where hundreds of models were fitted in order 
to find the one with the best parametrization) that a speed boost of around 40\% can be achieved.
For RB the models found even outperformed the MART models.

It could be interesting to further explore the core randomization process of RB and RF
and test different distributions from which tree depths are drawn. So far,
only the upper limit has been specified, while the lower bound has been fixed to a value of one.
One could examine versions where the lower tree depth bound can be chosen flexibly as well.
Moreover, one could also change the overall probability distribution of the tree depth parameter
to put more probability mass on depths that seem more plausible. This way,
prior beliefs about the interaction depth governing the data system could be incorporated.

Finally, in future work we will investigate the new fitting approaches in a more extensive 
simulation study with different covariate settings and covariate effect types as well
 as in complex real data applications.


\section*{Appendix}

\IncMargin{1em}
\begin{algorithm}[H] 
\DontPrintSemicolon

\SetKwInOut{Input}{input} \SetKwInOut{Output}{output}
\SetKwProg{Fun}{Function}{}{}
\vspace*{0.3cm}

\Input{Training data $\Lset$ of dimensions $N \times p$, number of randomly selected variables for each split $m_{try} \le p$ and number of iterations $B \in \mathbb{N}$}
\Output{A Random Forest object consisting of an ensemble of trees $\{ f_b \}_{1}^{B}$}
\BlankLine

\For{$b = 1$ \KwTo $B$}{
	Draw a bootstrap sample $\Bset_b$ of size N from $\Lset$\;
	Grow the RF tree $f_b(\cdot)$ on $\Bset_b$ by recursive partitioning so long as the number of observations is above threshold $n_{min}$ \;
	\ForEach{internal node $I$ in tree}{
		\eIf{node size is below threshold value $n_{min}$}{
			\textbf{stop}\;
		}{
			Select $m_{try} \le p$ variables at random\;
			Find the best (attribute, value) split point $(x_j, s^*)$\;
			Split $I$ into two children on $x_j$ using $s^*$\;
		}
	}
}
Combine the trees: $\hat{f}_{rf}(\x) = \frac{1}{B} \sum_{b = 1}^{B}\hat{f}_b(\x)$\;
\vspace*{0.3cm}

\caption{Random Forest in regression} \label{algo:random_forest}
\end{algorithm} \DecMargin{1em}

\IncMargin{1em}
\begin{algorithm}[H] 
\DontPrintSemicolon

\SetKwInOut{Input}{input} \SetKwInOut{Output}{output}
\SetKwProg{Fun}{Function}{}{}
\vspace*{0.3cm}

\Input{Training data $\Lset$ of dimensions $N \times p$,  and number of iterations $M \in \mathbb{N}$}
\Output{A boosting object consisting of an ensemble of learners $\{ f_m \}_{m = 1}^{M}$}
\BlankLine

set $\w_0 = (\frac{1}{N}, \frac{1}{N}, \dots,  \frac{1}{N})^\top \in \mathbb{R}^N$\;
\For{$m = 1$ \KwTo $M$}{
	use $\Lset$ weighted by $\w_{m-1}$ and learn base procedure $f_m(\cdot)$\;
	compute the weighted in-sample misclassification rate \;
	\[
		err_m = \frac{ \sum_{i = 1}^{N} w_{(m-1)i} \1(y_i \neq f_m(\x_i)) }{ \sum_{i = 1}^{N} w_{(m-1)i} }
	\]\;
	compute the weight the base classifier receives in final vote \;
	\[
		\alpha_m = \log \frac{1 - err_m}{err_m}
	\]\;
	update weights \;
	\[
		w_{mi} \leftarrow w_{(m-1)i} * \exp(\alpha_m * \1(y_i \neq f_m(\x_i)), i = 1, 2, \dots, N
	\] \;
}
Construct the estimate: $\hat{f}_{AdaBoost}(\x) = \argmax_{y \in \lbrace0,1\rbrace} \sum_{m=1}^{M} \alpha_m \1(\hat{f}_m(\x) = y)$\;
\vspace*{0.3cm}

\caption[AdaBoost]{A description of AdaBoost, see \citet{freund1996experiments} for the originial.} \label{algo_adaboost}
\end{algorithm} \DecMargin{1em}
\vspace{1cm}

\IncMargin{1em}
\begin{algorithm}[H]  \label{algo:ls_boost}
\DontPrintSemicolon

\SetKwInOut{Input}{input} \SetKwInOut{Output}{output}
\SetKwProg{Fun}{Function}{}{}
\vspace*{0.3cm}

\Input{Training data $\Lset$ of dimensions $N \times p$,  base learner class (e.g.\ regression tree), and number of iterations $M \in \mathbb{N}$}
\Output{A boosting object consisting of an ensemble of learners $\{ f_m \}_{0}^{M}$}
\BlankLine

set $\hat{F}_0(\x) = \hat{f}_0(\x) = \mean{y}$\;
\For{$m = 1$ \KwTo $M$}{
	Compute residuals $\tilde{y}_i = y_i - \hat{F}_{m-1}(\x_i), i=1,\dots,N$ \;
	Fit the residuals with least-squares  and obtain $\beta_m$ \;
	Carry out line search and get $\alpha_m = \rho_m$ \;
	Update model: $F_m(\x) = F_{m-1}(\x)  + f_m =  F_{m-1} + \alpha_mf(\x;\bm{\beta}_m)$
}
Construct the estimate: $\hat{F}_M(\x) = \sum_{m = 0}^{M}\hat{f}_m$\;
\vspace*{0.3cm}

\caption[Gradient boosting with least-squares -- The Multiple Additive Regression Trees (MART) Algorithm]{Gradient boosting with least-squares, see \citet{friedman2001greedy} for the originial.} \label{algo:ls_boost}
\end{algorithm} \DecMargin{1em}

\pagebreak

\bibliographystyle{chicago}
\bibliography{bib/references}


\end{document}